\def\year{2020}\relax
\documentclass[letterpaper]{article} 
\usepackage{aaai20}  
\usepackage{times}  
\usepackage{helvet} 
\usepackage{courier}  
\usepackage[hyphens]{url}  
\usepackage{graphicx} 
\usepackage{graphicx}  
\usepackage{times}
\usepackage{xcolor}
\usepackage{soul}
\usepackage[utf8]{inputenc}
\usepackage[small]{caption}
\usepackage{amssymb}
\usepackage{graphicx}
\usepackage{stmaryrd}
\usepackage{booktabs}
\usepackage{array}
\usepackage{amsmath}
\usepackage{nicefrac}

\usepackage[utf8]{inputenc} 
\usepackage[T1]{fontenc}    
\usepackage{url}            
\usepackage{amsfonts}       
\usepackage{nicefrac}       
\usepackage{microtype}      
\usepackage{amsthm}
\usepackage{amsmath}
\usepackage{mathtools}
\usepackage[noend]{algpseudocode} 

\usepackage{graphicx}
\usepackage[font=small]{subcaption}

\newcommand{\citet}[1]{\citeauthor{#1}~\shortcite{#1}}
\newcommand{\citep}{\cite}

\newcolumntype{x}[1]{>{\centering\arraybackslash\hspace{0pt}}p{#1}}
\usepackage{color}
\usepackage[textsize=scriptsize]{todonotes}
\definecolor{blue}{RGB}{0, 93, 170}			

\usepackage{xspace}
\newcommand{\dgem}{\text{\textsc{MCN-GEM}}\xspace}

\title{A Multi-Channel Neural Graphical Event Model with Negative Evidence}

\author{
 Tian Gao${^\dagger}{^{*}}$, Dharmashankar Subramanian$^\dagger$\thanks{ indicates equal contribution}, Karthikeyan Shanmugam$^\dagger$, 
\\ 
{\Large \bf Debarun Bhattacharjya$^\dagger$, Nicholas Mattei$^\S$ }
\\
$^\dagger$ Research AI, IBM T. J. Watson Research Center, Yorktown Heights, NY, USA \\ 
$^\S$ Department of Computer Science, Tulane University, New Orleans, LA, USA \\
\{tgao, dharmash, debarunb\}@us.ibm.com\\ Karthikeyan.Shanmugam2@ibm.com, nsmattei@tulane.edu  
}

\begin{document}

\maketitle

\begin{abstract}
Event datasets are sequences of events of various types occurring irregularly over the time-line, and they are increasingly prevalent in numerous domains. 
Existing work for modeling events using conditional intensities rely on either using some underlying parametric form to capture historical dependencies, or on non-parametric models that focus primarily on tasks such as prediction. We propose a non-parametric deep neural network approach in order to estimate the underlying intensity functions. We use a novel multi-channel RNN that optimally reinforces the negative evidence of no observable events with the introduction of fake event epochs within each consecutive inter-event interval. We evaluate our method against state-of-the-art baselines on model fitting tasks as gauged by log-likelihood. Through experiments on both synthetic and real-world datasets,  we find that our proposed approach outperforms existing baselines on most of the datasets studied.
\end{abstract}

\section{Introduction}
Event stream data is collected to explore the dynamics and behavior of a wide variety of systems including social networks, biochemical networks, electronic health records, and computer logs in data centers. In a multivariate event stream, events of multiple types (labels) occur at irregularly spaced time stamps on a common timeline. Event models seek to capture the joint stochastic dynamics of such event streams. Multivariate point processes and \emph{conditional intensity functions} provide the mathematical framework for capturing event dynamics \citep{aalen2008}. In this framework, the instantaneous arrival rate of an event type at any point in time typically depends on the history of all historical event arrivals before that point in time.

The model fitting task for event stream data has a long history in machine learning, including prior work in  temporal point process modeling \cite{gunawardana2011,cpcim12,weiss2013,bsg2018}. It is fundamentally different from the prediction task as it is unsupervised since no ground truth (graph) model is given. Various models have been proposed in the literature to capture history-dependent arrival rates including graphical event models like proximal graphical event models \citep{bsg2018}, forest-based point processes \citep{weiss2013}, piecewise-constant conditional intensity models \citep{gunawardana2011}, and Poisson networks \citep{rajaram2005}, and others like non-homogeneous Poisson processes \citep{goulding2016}, continuous time noisy-or (CT-NOR) models \citep{simma2008}, and Poisson cascades \citep{simma2010}. One drawback is that all these approaches make assumptions about the parametric form of the corresponding model, and this is challenging in practice without first-hand knowledge of the underlying data generating process.

On the other hand, researchers have also proposed sequential deep learning techniques for event data sets \cite{xiao2017modeling} such as recurrent marked temporal point processes (RMTPP) \citep{du2016recurrent} and neural Hawkes processes \citep{mei2017neural}. These models use a recurrent neural network to capture the history dependency of the conditional intensity function. Note that these methods are semi-parametric; beyond just the network weights, they make functional assumptions for how to translate the hidden states of the network to a corresponding event arrival rate. They also show a remarkable similarity to the Hawkes process where each historical event exerts an additive influence on conditional intensity, and which decays over time. In neural Hawkes \citep{mei2017neural}, the cell state of the network decays exponentially. In RMTPP \citep{du2016recurrent}, there is a single event arrival rate whose epochs carry a label to capture multiple event types, and this single rate is an exponential function of the sum of a baseline intensity along with additive influences from the past. While such representations can be powerful and expressive, they may not adequately learn general-purpose forms of history-dependence. For example, consider history-dependence that doesn't involve exponential decay; examples include piece-wise constant intensities 
\cite{gunawardana2011,bsg2018}, 
or processes involving time lags with delayed excitation or inhibition, or if different event types follow varying time scales, etc. Non-simple-decay and delayed, nonlinear intensity functions are not satisfactorily addressed by existing works.

In this paper, we propose a non-parametric deep learning approach to model multivariate event data sets in continuous-time. We seek to learn history-dependent conditional intensity functions in a fully data-driven, non-parametric manner, i.e. using only network weights and activation functions, and via learning a suitable representation of all available (strict) histories. Firstly, we note that the inter-event interval is an important source of information for modeling event data, i.e. not only is the presence of an event at a point in continuous time an important signal for learning, but also the absence of an event at a point in time. We present a deep learning model that reinforces this negative evidence. We also note that real-life events exhibit multiple time scales. Historical sequences of fixed length that are used in classical recurrent neural networks (RNNs) do not suffice to address this issue. We present a multi-scale, multi-channel sequential representation that is sensitive to the base rate of various event types. Lastly, real-life data sets exhibit complex history dependencies like delayed excitation due to events of either the same or another type. We use different types of spatial (across types) and temporal (across time) attention models. 
Our work is motivated by the framework of graphical event models (GEMs) \citep{gunawardana2016} where the historical event epochs of a set of parents affect the instantaneous rate of a child point process.

\paragraph{Contribution.} We make the following contributions: (1) We propose a simple, yet effective, non-parametric way to approximately capture the continuous-time variation of historical influence on conditional intensity by exploiting the negative evidence from each successive inter-event duration. (2) We develop an efficient multi-scale, multi-channel internal state representation, which lets us align our architecture with graphical event models. (3) We also propose a spatio-temporal attention model to capture whose histories (among event labels), and which points in time are most influential toward determining the instantaneous arrival rate for any chosen label. The resulting proposed {\bf{multi-channel neural graphical event model (\dgem)}} is thereby similar to structure learning in the graphical modeling literature. We demonstrate non-trivial gains in evaluating point process log likelihood estimates on test data over state-of-the-art models.

\section{Multi-Channel Neural Graphical Event Model}


\begin{figure}[ht!]
\centering
\begin{subfigure}{1.0\linewidth}
	\centering
	\includegraphics[width=.95\columnwidth, page=1]{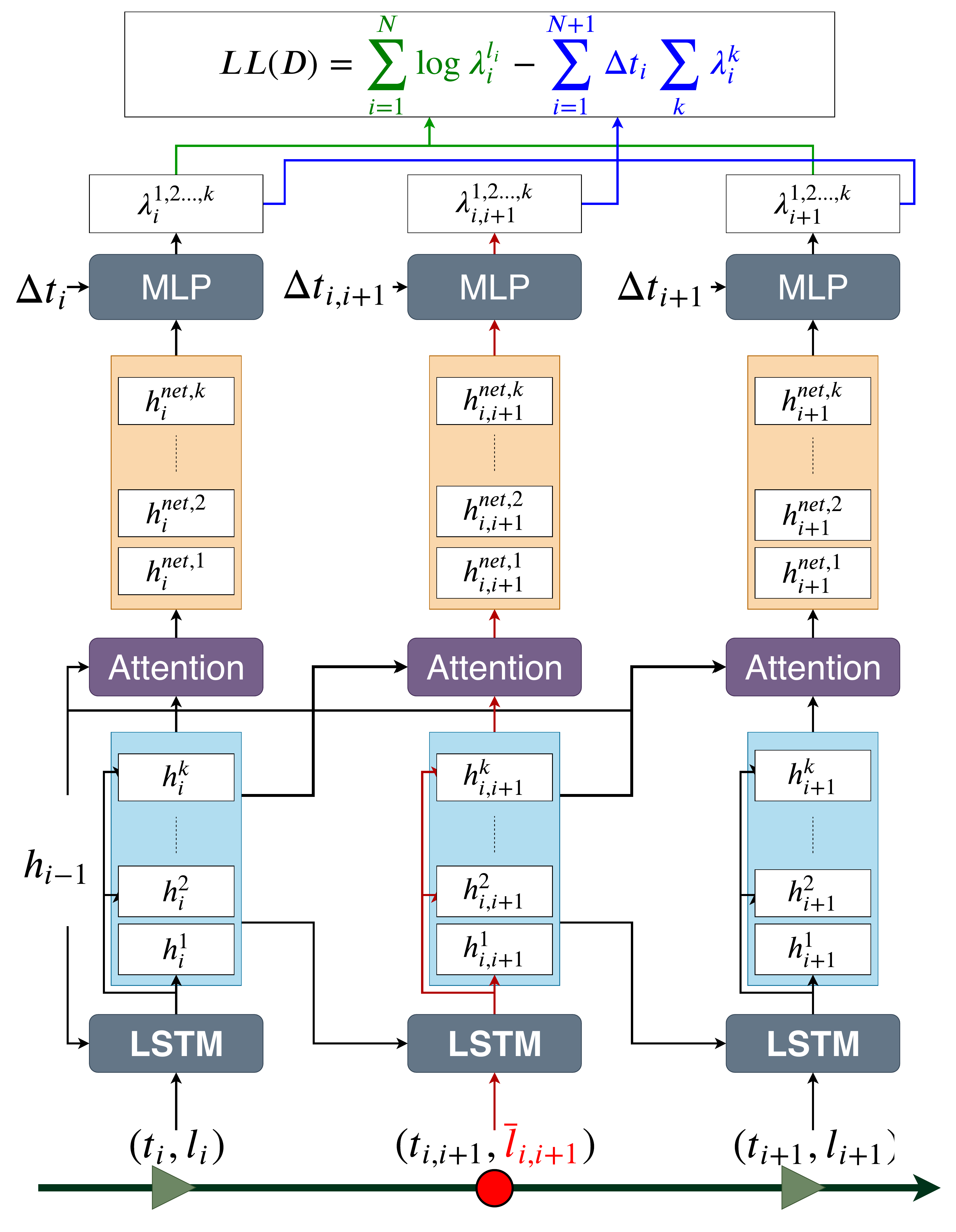}

\end{subfigure}
\hfill
\caption{System overview of the proposed MCN-GEM model with negative evidence $\bar{l}$.}
\label{fig:system}
\end{figure}

Following the notation in graphical event models \citep{gunawardana2016,bsg2018}, we denote an event dataset as $D = \{ (l_i , t_i) \}_{i=1}^N$, where $t_i$ is the occurrence time of the $i^{th}$ event, $t_i \in \mathbb{R}^+$, and $l_i$ is an event label corresponding a finite set $\mathcal{L}$ of possible labels (types), where $|\mathcal{L}| = M$. 
We assume a time ordered dataset where at most one event may occur at any point in continuous time. So we have $t_i < t_j$ for $i < j$, with initial time $t_0 = 0 \leq t_1$ and end time $t_{N+1} = T \geq t_N$, where $T$ is the total time period. We denote the strict history relative to any point $t$ in continuous time as $\mathcal{H}_t$, and this is defined as the sequence of event epochs before time $t$, i.e. $\{(t_i, l_i) | t_i < t\}$. Lastly, $\lambda^{k}_{t}|\mathcal{H}_t$ denotes the instantaneous conditional intensity of event type $k$ in label set $\mathcal{L}$ at time $t$. It governs the history-dependent instantaneous rate of occurrence of event type $k$ at time $t$. Note that we use the term epoch to denote an event epoch, i.e. an event arrival on the timeline. Figure~\ref{fig:system} shows the overall system of our proposed MCN-GEM, which consists of three main components: a recurrent neural network to represent the continuous time evolution, the introduction of negative evidence between two consecutive arrival epochs, and the multi-channel view with spatial and temporal attention. Below we discuss each component in detail.

\subsection{Continuous Time in Deep Event Models}
Given a dataset $D$, a  graphical event model (GEM) $\mathcal{G}$ is a directed graph with nodes $\mathcal{L}$ and edges $\mathcal{E}$. It defines a family of marked point processes whose likelihood of data $D$ is \citep{gunawardana2016}:

$$p(D|\theta) = \sum_{l\in \mathcal{L}} \sum_{i=1}^n \lambda_{t_i}^{l_i \cdot \textbf{1}_{l_i}}  \cdot e^{\int_{t_{i-1}}^{t_{i}} \sum_{k=1}^{M} \lambda^{k}_{\tau}|\mathcal{H}_{\tau} d \tau} $$

where the indicator function $\textbf{1}_{l_i} = 1$ if $l_i=l$ and zero otherwise.  

One may write the log-likelihood of the event data over interval $[0,T]$ using the conditional intensity functions as:
\begin{equation}
\text{logL}({D}) = \sum_{i=1}^{N} \log \lambda^{l_i}_{t_i}|\mathcal{H}_{t_i} - \sum_{i=1}^{N+1} \int_{t_{i-1}}^{t_{i}} \sum_{k=1}^{M} \lambda^{k}_{\tau}|\mathcal{H}_{\tau} d \tau
\label{eqn:log-likelihood}
\end{equation}
Our objective is to train a deep neural network to produce $\lambda^{k}_{t}|\mathcal{H}_{t} $ for each type $k$. We use a sequence modeling approach with recurrent neural networks and long short term memory (LSTM) cells. The sequence of tokens that we feed into the LSTM network is the temporally ordered event sequence $D$, where each token corresponds to an event arrival, i.e., a label and a time stamp. To model continuous time, each token is represented in the raw input as a concatenation of its one-hot encoded event label (1-out-of-$M$) and its continuous-valued time stamp, similar to previous work \citep{du2016recurrent,mei2017neural}. The internal states of the recurrent LSTM cell evolve in response to each current raw input. We note that there are alternative choices for modeling continuous time, such as contextualizing the event label embedding with a time mask \citep{li2017time}. 

 For an event sequence $D$ as per our notation, $l_0$ denotes a token that marks the beginning of the sequence and $l_{N+1}$ denotes a token that marks the end of the sequence. Then for each event epoch $i$ in sequence $D$, we have $0 \leq i < N+1$, and we compute the hidden state $h_{i}^k$ for each event label $k$ in the augmented set $\mathcal{L}$ as:

$$h_{i}^k= \text{LSTM}_k( [\text{Emb}(l_i), t_i]; h_{i-1}^k), \quad  \forall k \in \mathcal{L}$$

where $\text{Emb}$ denotes the embedding matrix for label $l_i$. Embedding consists of one embedding layer on top of one-hot encoding of labels. $h_{-1}^k$ is initialized to be all-zero vectors. 
For practical computation reasons, we find that sharing the LSTM parameters among different event labels $k$ is sufficient, i.e., $\text{LSTM}_k = \text{LSTM}$. 

\subsubsection{Basic Model.} 
\begin{figure*}[ht!]
\centering
\begin{subfigure}{0.75\linewidth}
	\centering
	\includegraphics[width=\linewidth, page=1]{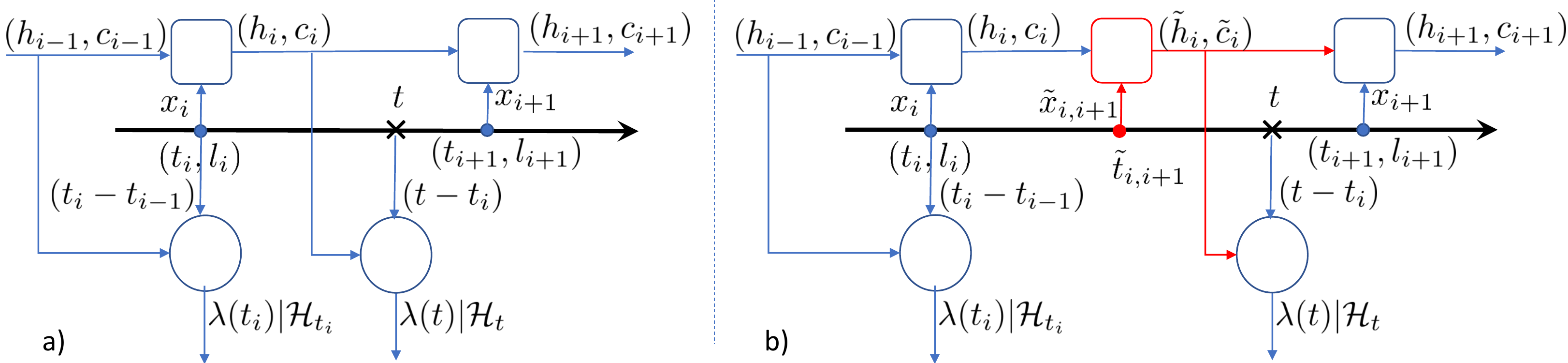}

\end{subfigure}
\hfill

\caption{Deep sequential model for conditional intensity: a) A basic model, b) A model with fake event epochs to reinforce negative evidence. }
\label{fig:rnn_basic}
\end{figure*}

Figure \ref{fig:rnn_basic} (a) shows a basic model for history dependent conditional intensity. The black lines show the common timeline on which two consecutive epochs $(t_i, l_i)$ are shown. The upper half shows the recurrent cells unrolled in time,and the bottom half shows a neural network layer, namely the $\lambda$-network, to produce the instantaneous conditional intensity vector $\lambda$. The recurrent cell receives two inputs at each sequential token (event arrival): the raw input $x_i$ which is the concatenation of the one-hot encoded event label $l_i$ and the time stamp $t_i$, and the penultimate value of the internal states ($h_{i-1}, c_{i-1})$. The sub-network that produces the conditional intensity ($\lambda(t) | \mathcal{H}_t$) takes two inputs: the most recent internal states that are strictly before $t$, i.e. in correspondence with the most recent event (say at $t_i < t$), and the duration $(t-t_i)$. 

\subsection{Modeling Continuous Time History with Negative Evidence}

\subsubsection{Motivation on New Interval State Representation.} We note that the internal states ($h_i, c_i$) in the basic model evolve discretely across tokens while staying fixed in between tokens. In event modeling, the inter-event duration between any two consecutive events is just as important as the event arrivals themselves, as evident from the integral terms in Equation \ref{eqn:log-likelihood}. As such, it is important that the internal states of the sequential model also reflect the continuous passage of time in between successive events. The neural Hawkes process \citep{mei2017neural}  proposes capturing this continuous variation using an exponential decay function, motivated by the classical Hawkes process \citep{hawkes1971}. However, this raises the question of whether an exponential form of decay is always the best choice, and whether there is a more non-parametric alternative that adapts to data. For example, consider proximal graphical event models \citep{bsg2018} and the more general family of piece-wise constant intensity models \citep{gunawardana2011} that don't exhibit exponential decay. We propose the use of "fake epochs" to reinforce the negative evidence of no observable events within each inter-event dead-space as shown in Figure  \ref{fig:rnn_basic}(b).

\subsubsection{Fake Epochs as Negative Evidence.} We introduce an auxiliary $(M+1)^{\text{th}}$ label into our label set $\mathcal{L}$, and refer to it as the "fake label". We then interject within each inter-event interval a certain number $K$ of fake epochs (i.e. with label $M+1$) that are spread uniformly in time over the interval. Note that $K$ is a hyperparameter, and Figure \ref{fig:rnn_basic}(b) shows one fake epoch ($K=1$) introduced at $\tilde{t}_{i,i+1}$, which is the center of $[t_i, t_{i+1}]$. The fake event epoch then participates in the recurrent computations like any other real event epoch. It allows using the LSTM dynamics to further evolve the internal states  within each inter-event interval, albeit in a discrete manner i.e. at each such fake event epoch on the timeline. The resulting finer sequence of internal states is then a summary of all the event trace history as well as the passage of time in the intervening dead-space intervals. Note how the internal state has evolved to $(\tilde{h}_i, \tilde{c}_i)$ in the interior of the interval. The fake event epochs also allow us to compute the integral terms in Equation \ref{eqn:log-likelihood}. We take a numerical quadrature procedure using the fake epoch time stamps as the sampling time points for the quadrature. We note that while the entire inter-event interval corresponds to negative evidence in theory, the proposed finite approximation suffices in practice as shown later in the experimental section. Lastly, while the conditional intensity corresponding to the fake label doesn't enter Equation \ref{eqn:log-likelihood} in any way, we use it to regularize the learning objective via a target-label reconstruction term described later.

A similar intent of using fake epochs was pursued in a different graphical model \citep{Gopalratnam:2005:ECT:1619410.1619490}, which proposed to model the negative evidence of no observable events. It also has some similarity with generative adversarial training \cite{ganin2016domain,xiao2017wasserstein}. The fake epochs with a fake event label are akin to adversarial sample points relative to the real set of labels. Using the fake epochs to regularize the learning via target label reconstruction, or via binary classification of real-versus-fake makes it similar to adversarial training.


\subsection{A Multi-Channel View for Spatial and Temporal Attention}


In multivariate event data, the different labels may  have different arrival rates. Further, they may mutually influence the arrival rates of each other in label-specific ways in the sense of graphical event models. It is therefore desirable to model a label-specific hidden state. We achieve this by associating each label with a corresponding partition of a single hidden state vector. We achieve this by choosing the hidden state dimension to be an integer multiple of the number of labels, i.e. $m(M+1)$ for some positive integer $m$, and effectively realize an $m$-dimensional hidden state for each label in $\mathcal{L}$. These label-specific sub-vectors are selectively channeled for computing the label-specific rates through the network layer in Figure \ref{fig:rnn_basic}. Such a multi-channel view also enables modeling temporal and spatial attention within the $\lambda$-network.

\begin{figure}[t]
\centering
\includegraphics[width=1\linewidth]{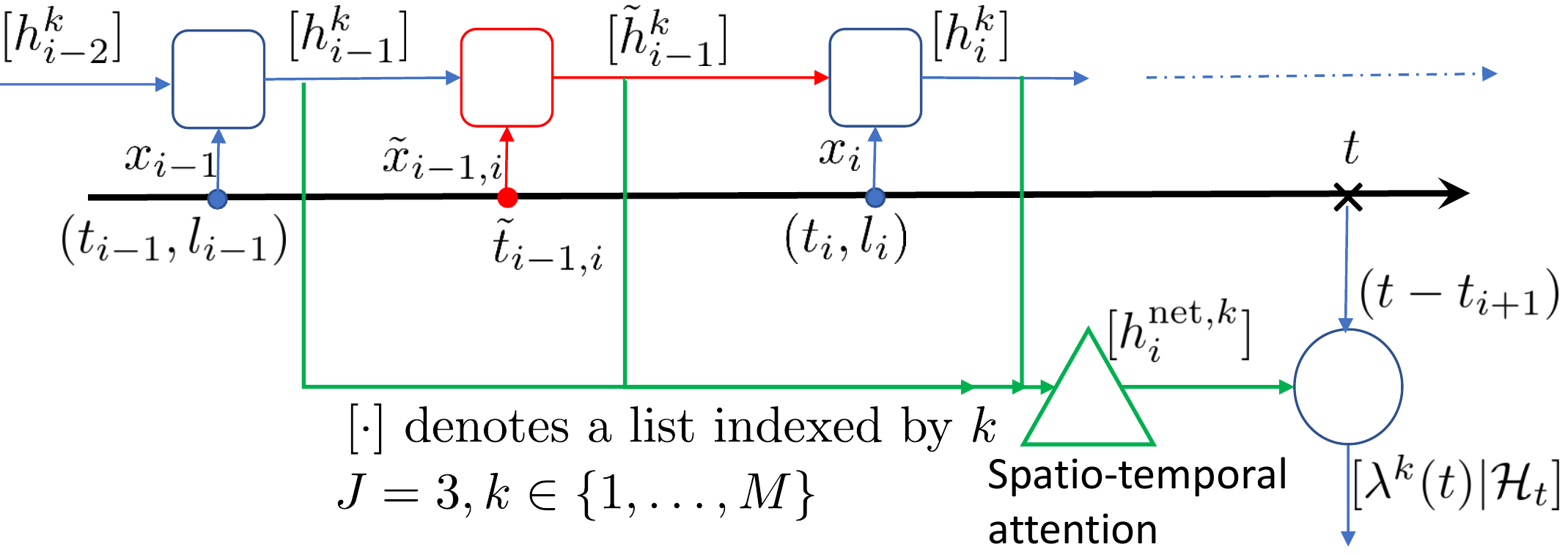}
\caption{Spatio-temporal attention for deep graphical event models.}
\label{fig:spatiotemporal}
\end{figure}

Spatial attention is a way to model inter-label dependence, as in graphical event models, and temporal attention is a way to model the lagged dependence on parental event history. For example, in a piece-wise constant intensity model \citep{gunawardana2011}, the active basis functions consist of lagged intervals relative to any point in time.  We achieve these two types of dependence by maintaining a memory bank $\mathcal{M}$ of historical label-specific hidden states that span the most recent $J$ event arrivals. Note that the above multi-channel view gives us a total of $M*J$ hidden state sub-vectors in $\mathcal{M}$, i.e. one hidden-state sub-vector for each real label and for each time-stamp corresponding to the most recent $J$ event arrivals. These raw hidden states in the memory bank are combined using an attention mechanism into a net hidden state that becomes input for the lambda-network. Figure \ref{fig:spatiotemporal} shows this for $J = 3$, where $[h^{k}_{i}]$ in the figure denotes a list of hidden states indexed by label $k$, and at time $t_i$. The attention block produces a corresponding list of net hidden states $[h^{\text{net},k}_i]$  which enter the $\lambda$-network to produce conditional intensities for each label at time $t$.

By adding an attention layer after $h_i^k$, intensity rates can be shown to credit the relevant event occurrences in history. Our memory bank $\mathcal{M}_i$ (for each epoch $i$) has size $J \times M$ to stores the raw hidden states of each of $M$ channels from the most recent $J$ epochs relative to $i$, i.e. from $(i-J)$ till $(i-1)$. Note that $J$ is a constant hyperparameter in the model. The net attentive hidden state is given as,

$${h}_{i}^{\text{net},k}= \text{tanh} ( W_c[c_i^k, h_{i}^k]), \quad  \forall k \in \mathcal{L}$$

where $c_i$ is the context vector at epoch $i$, and is computed as a weighted average of the raw hidden states in $\mathcal{M}_i$  The weighting is done by an alignment vector $\alpha^k_i$, i.e., $c_i^k = \sum_m \alpha_{im}^k h_{m}$, where $h_{m}$ is each raw label-specific hidden state in the memory bank  $\mathcal{M}_i$ Note that index $m$ runs over each of the $JM$ hidden states in  $\mathcal{M}_i$ thereby giving both temporal and spatial attention. $W_C$ represents weights to compute new hidden representation given attention.  The attention weight  $\alpha_{im}^k$ is derived by comparing the current raw hidden state $h_{i}^k$ to each raw hidden state $h_{m}$ in the memory bank at time $i$:

$$ \alpha_{im}^k = \text{align}( h_{i}^k, h_{m}) = \frac{exp(score( h_{i}^k, h_{m})}{\sum_m exp(score( h_{i}^k, h_{m}))} $$

where $score( h_{i}^k, h_{m}) = (h_{i}^k)^T h_{m}$. We note that other types of attentions can be used as well in the above spatio-temporal attention model.

\subsection{Training}

We then use two feed-forward  layers to learn intensity rate $\lambda_{t_{i+1}}^k$ given the net hidden state ${h}_{i}^{\text{net},k}$ and time interval $\Delta t_{i+1} = t_{i+1} - t_{i}$:

$$\lambda_{t_{i+1}}^k  |\mathcal{H}_{t_i}=  
\sigma_2 (f_2( \sigma_1( f_1( [{h}_{i}^{\text{net},k}, \Delta t_{i+1}]  ) )), \quad \forall k \in \mathcal{L}$$

where $f_1$ and $f_2$ are feed-forward neural layers. We use a ReLU and softplus activation function for $ \sigma_1(\cdot)$ and $\sigma_2(\cdot)$, respectively. The Softplus activation ensures a positive conditional intensity, which is a standard component and has been used in previous deep models \cite{mei2017neural}.

To train \dgem, we use the same LL function as in Equation~\ref{eqn:log-likelihood} with the assumption of constant intensity in between two consecutive events, real or fake, and this leads to:

\begin{equation}
LL(D) = \sum_{i=1}^{N} \log \lambda^{l_i}_{t_i}|\mathcal{H}_{t_i} - \sum_{i=1}^{N+1} \Delta t_i \sum_{k=1}^{M} \lambda^{k}_{t_i}|\mathcal{H}_{t_i}
\label{eqn:obj}
\end{equation}

where $\Delta t_i = t_i - t_{i-1}$ is the time interval since the last event.

In addition, since our model is the most non-parametric compared to other related work, we find that adding two regularization terms helps with generalization. First, we consider the target prediction loss $\mathcal{L}_p$ of the next event label $l_{i+1}$ given $\lambda_{t_{i+1}}^k$, for which we use a cross entropy loss between ground truth $l_{i+1} $ and softmax$(\lambda_{t_{i+1}})$. This classification loss also emphasizes the adversarial nature between real and fake epochs. Second, we add another term $\mathcal{L}_w$ which penalizes the $\mathcal{L}_2$-norm of the weights on $f_1$ and $f_2$. Hence, the overall regularized objective for training is as follows:

\begin{equation}
\mathcal{L}_{\text{train}} = LL(D) + \lambda_p \mathcal{L}_p + \lambda_w \mathcal{L}_w 
\label{eqn:obj_train}
\end{equation}

For the model fitting computation task, the overall likelihood objective for training is computed as Equation~\ref{eqn:obj_train}. The testing loss is simply computed as $\text{LL}(D)$, since our aim is to gauge how well the model fits the data.




\section{Empirical Evaluation}
\label{sec:exp}
We compare our methods with state-of-the-art algorithms for model fitting of event streams. 
We compare our proposed neural method with a recent model in the GEM class, PGEM \cite{bsg2018}, and the most relevant neural baseline  Neural Hawkes Process (NHP) \cite{mei2017neural}. Some other works \cite{xiao2017modeling,du2016recurrent} also use neural networks but do not perform the model fitting task. We implement \dgem in Pytorch, PGEM in Python, and use publicly available code for NHP\footnote{https://github.com/HMEIatJHU/neurawkes}
. Please refer to the appendix for further implementation details of \dgem, especially on choices of hyper parameters. Our proposed MCN-noF refers to the version of our model where no fake epochs are introduced; we show this to highlight the effect of fake epochs. If not otherwise specified, \dgem uses one fake epoch. We use log likelihood as the metric to evaluate model fitting and bold the best performing approach.  

For our experiments, we divide an event dataset into $70\%-30\%$ train-test splits. For data involving single event streams, we perform the split based on time, i.e. select events in the first $70\%$ of the entire duration in the train set. For data involving multiple streams, we split by randomly selecting a subset of $70\%$ of the streams in the train set. All shown results are those evaluated on the test set. All methods are compared in the same experimental setting. 

\paragraph{Synthetic Datasets. }
We first conduct experiments using event streams based on proximal graphical event models (PGEMs). The data is generated following the same sampling procedure described in \citet{bsg2018}. Specifically, for each node, we first sample the number of its parents $K$,  the set of parents, windows, and then intensity rates in that order. Please refer to the supplement material for exact details and generating parameters. 

We generate $5$ PGEMs, each with $5$ nodes. We use default values to generate the models as provided in the supplementary material of \citet{bsg2018}. For each model, $10$ event streams are produced with a synthetic PGEM data generator, up to $T=1000$. Table~\ref{table:synth} shows the performance of the algorithms on the test sets for the $5$ PGEM models. We observe that when PGEM is the data-generating model, all methods including neural-based models can recover a log likelihood (LL) very close to the PGEM learner. What is particularly striking is that with the introduction of a fake epoch, \dgem performs much better than the other models, even the PGEM learner, usually by $20\%$ to $35\%$. We note that the state-of-the-art PGEM learner \citep{bsg2018} involves a greedy coordinate ascent approach for learning the optimal windows and graphical structure, which allows for improvement using our non-parametric neural model with a limited number of data. The quadrature approximation that we use (see Equation \ref{eqn:obj_train}) across the augmented set of consecutive inter-event intervals is effectively a fine piecewise-constant conditional intensity model, thereby giving \dgem enough expressive power to effectively approximate the true generating PGEM.


\begin{table}
\centering
\centering
\resizebox{1.0\linewidth}{!}{
    \begin{tabular}{cccccccc}
        \toprule
    Dataset     &  PGEM & NHP & MCN-noF & \dgem \\
         \midrule
PGEM1 &  -6207.3 &-6248.8  & -6268.1  & \textbf{-4764.2}    \\
PGEM2 &  -8285.4 &-8289.4 & -8292.0  & \textbf{-5739.2} &   \\
PGEM3 &  -7933.1 &-8023.0 &-8177.7  & \textbf{-5714.8}   &  \\
PGEM4 &  -8853.2 &-8895.7 &-8904.3  & \textbf{-5608.5}  &   \\
PGEM5 &  -7811.0 &-7842.5 &-7912.5  & \textbf{-5601.6}  &  \\
         \bottomrule
    \end{tabular}
}
\caption{Average LL score for Synthetic Datasets.}
\label{table:synth}
\hfill
\centering

\resizebox{1.0\linewidth}{!}{
    \begin{tabular}{cccccccc}
        \toprule
    Dataset     &  PGEM & NHP &  MCN-noF & \dgem \\
         \midrule
Argentina & -3150.6  & -3181.9 & -3559.5 & \textbf{-2677.9} &  \\ 
Brazil & -3865.7  & -3822.3 &-4242.3& \textbf{-3329.0}  &  \\
Venezuela & -1663.2 & \textbf{-1567.4} & -1915.9 &-1655.1   &  \\
Colombia & -1095.1  & -1128.5&-1244.5 & \textbf{-1029.3}   &  \\
Mexico  & -1927.6  & -1936.5 &  -2375.7 & \textbf{-1756.7}   &  \\

\bottomrule
\end{tabular}
}
\caption{Average LL score for ICEWS Datasets.}
\label{table:icews}
\end{table}

\paragraph{Political News Datasets.}

A real world example of numerous, asynchronous events on a timeline are socio-political world events.  We use the Integrated Crisis and Early Warning System (ICEWS) political event dataset \cite{obrien2010} to test our model. This dataset is a set of dyadic events, i.e., \emph{X does Y to Z}, encoded as a set of over 100 actors and 20 high level actions from the Conflict and Mediation Event Observations (CAMEO) ontology \cite{GSYA02a}. The events in ICEWS take the form of ``The Police (Brazil) has Verbal Conflict with Protesters (Brazil).''  Following the same pre-processing as \citet{bsg2018}, we restrict our attention to a 4 year time period from Jan 1, 2012 to Dec 31, 2015 and use 5 countries, 5 actor types, and 5 types of actions.

Table~\ref{table:icews} compares model fit on these datasets across methods. Performances of PGEM and NHP are similar on these datasets, and \dgem with no fake epochs performs worse than NHP. It is possibly harder to learn without some parametric grounding in these datasets. However, with the introduction of fake epochs, \dgem again performs better than the other models by $5\%$ to $15\%$, with the exception of Venezuela, on which NHP performs the best.
For Venezuela, we suspect that MCN-GEM's proficiency in learning from history is slightly detrimental as there are discrepancies between train and test sets; the data exhibits a significant increase in event occurrences in recent years (test set) compared to previous years (train set). 

\paragraph{Healthcare Dataset. }
Following \citet{mei2017neural}, we also test algorithms on the MIMIC-II dataset, which includes electronic health
records from various patients with clinical visit records in Intensive Care Unit for 7 years. Each patient has a sequence of hospital visit events, and each event records
its time stamp and diagnosis, which serves as the event label. 

Since the PGEM learner cannot handle datasets with unknown labels during testing time, we also create a clean version of MIMIC, MIMIC-short (MIMIC-s), ensuring that MIMIC-s train and test have a common label set. We compare our method with PGEM on MIMIC-s, and with NHP on the original MIMIC. Table~\ref{table:mimic} shows again that \dgem with the fake epochs outperforms both NHP and PGEM by $15 \sim 20\%$ for the log likelihood computation. 

\begin{table}[!ht]
\centering
\resizebox{0.90\linewidth}{!}{
    \begin{tabular}{c|ccc|}
        \toprule
          & NHP &  MCN-noF & \dgem  \\

         \midrule
MIMIC   &-1165.0 & -1380.4 & \textbf{-837.4} \\
   \toprule
     & PGEM &  MCN-noF & \dgem \\
            \midrule
MIMIC-s & -677.8 &-704.4  & \textbf{-586.8}  \\

         \bottomrule
    \end{tabular}
}
\caption{Log likelihood for MIMIC Dataset.}
\label{table:mimic}
\end{table}

\subsection{Ablation Study} 

To gain more insights of the contribution of different components, we also study the effectiveness of each component of the proposed model via an ablation study, namely the effectiveness of multi-channel vs. single channel, the usage of attention vs. no attention, and the impact of attention memory bank sizes for negative evidence. As shown in Table~\ref{table:ablation} below (showing LL with standard deviation), we conducted an ablation study on different model components on ICEWS Argentina. We tested single channel (SC) vs. multi-channel (MC), one fake epoch (1F) vs. no fake (NoF), and different memory bank sizes (M=1,2,3,5,10). We use neural Hawkes (NHP) as the baseline. We repeat the experiments 5 times with 5 different random seeds, reporting their average $LL$ along with their standard deviation. 

\begin{table*}[!ht]\renewcommand{\arraystretch}{0.9}
\centering
    \begin{tabular}{c|cccccc}
        \toprule
         &NHP & SC noF & SC 1F & MC noF & MC 1F  \\
         \midrule

Argentina & $-3154.8\pm25.4$ & $-4323.8\pm35.4$  & $-3839.4.9\pm25.4$ & $-4282.2\pm58.1$ & $-3302.6\pm28.1$ &  \\
\toprule
& MC 1F 1M & MC 1F 2M & MC 1F 3M & MC 1F 5M & MC 1F 10M\\
  \midrule
Argentina & $-2841.0\pm23.5$ & $-2693.9\pm 27.0$ &$-2682.6\pm29.3$ &$-2871.8\pm126.9$ & $-2947.3 \pm 190.0$\\

         \bottomrule
    \end{tabular}
\caption{Ablation study of \dgem model components: LL performance on ICEWS Argentina.}
\label{table:ablation}
\end{table*}

\subsubsection{Impact of Multi-Channel Modeling.}  
As shown in the top row of Table~\ref{table:ablation}, the single-channel (SC noF and SC 1F) versions always perform worse than that of multi-channel versions (MC noF and MC 1F). It shows the importance of multi-channel modeling, by letting the hidden states evolved separately for each channel. 

\subsubsection{Impact of the Memory Bank Size.} 
As shown in the top row of Table~\ref{table:ablation}, without the attention (equivalent to the memory size of 0), the performance of the proposed model (MC-1F) is worse than that of NHP. The second row  of Table~\ref{table:ablation} shows the performance of MCN with memory size of 1  (1M), 2 (2M), 3(3M), 5(5M), and 10 (10M), and all outperform NHP. In addition, memory size of 3 is the best performing in Argentina, while the larger memory sizes could degrade the performance.

\subsection{Impact of the Number of Fake Epochs}
We study the impact of the number of fake epochs on the performance of \dgem through a study on the ICEWS Argentina dataset. We vary the number of fake epochs, from $0$ to $5$, to see whether adding more fake epochs would further improve the LL. We use 1F to indicate the usage of 1 fake epoch, 2F for 2 fake epochs, etc. As shown in Table \ref{table:numf}, introducing one fake epoch has a large gain in LL. Introducing many more leads to minimal improvements, and can even hurt the performance. 
We note that without fake epochs, the computation of integral in Equation~\ref{eqn:obj} is assumed to be piece-wise constant in each inter-event interval. This would perform badly in general, such as when there is decay, growth, or more complex temporal variations in conditional intensity. Even with 1 fake epoch, \dgem appears effective in approximating such variation. Introducing more fake epochs leads to a subtle trade-off. While they allow a finer approximation in the quadrature-sum for the second term, they also lead to an additional imbalance in the relative proportions of real and fake epochs. This may result in reduced intensity rates of real events at real epochs due to fitting their rates better at the fake epochs. Fake epochs are also related to the windows and basis functions in GEMs \citep{gunawardana2016}. Instead of manually providing a set of basis functions, \dgem uses the data to adaptively infer the start and expiration of new basis functions. It is very simple yet effective to do so in a non-parametric way.

\begin{table}\renewcommand{\arraystretch}{0.9}
\centering
\resizebox{1\linewidth}{!}{
    \begin{tabular}{cccccc}
        \toprule
          noF  & 1F &  2F & 3F &4F & 5F \\
         \midrule
-3559.5  &-2677.9 & \textbf{-2652.3}&  -2772.1 & -2803.1 &-2881.0 \\

         \bottomrule
    \end{tabular}
    }
\caption{Study of the number of fake epochs vs. LL performance on ICEWS Argentina.}
\label{table:numf}
\end{table}

\begin{figure}[t]
\centering
\includegraphics[width=0.48\textwidth]{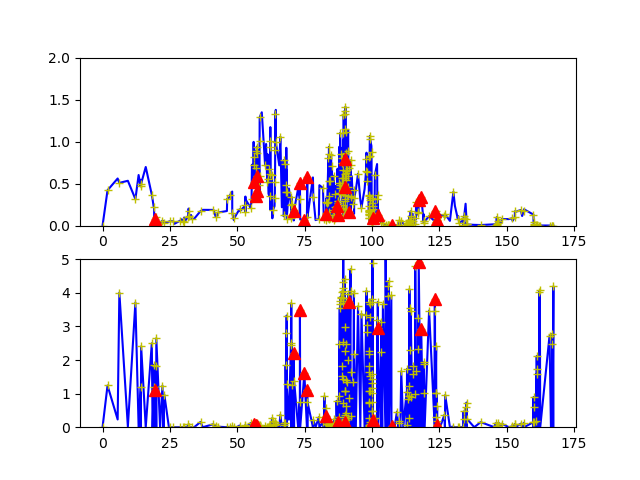}
\caption{Visualization of lambdas (conditional intensity rates) for Event Type 1 in ICEWS Argentina Dataset. The top shows rates from \dgem$-$noF, and the figure shows rates from \dgem.  Red triangles indicate the occurrence of Event Type 1, yellow crosses indicate the occurrence of all other real events. }
\label{fig:lambda}
\end{figure}

To visualize the impact of fake epochs on the learned intensity rates, we plot them from \dgem$-$noF and \dgem with one fake epoch in the ICEWS Argentina dataset. We show only Event Type 1's intensity rates in Figure~\ref{fig:lambda}, but the trends hold for all other event types. In Figure~\ref{fig:lambda}, the top figure shows the rates, in blue lines, of Event Type 1 from noF over time and the bottom figure shows rates from \dgem. A red triangle in both figures indicates the rate at the occurrence of Event Type 1, and a yellow cross indicates the rate at the occurrence of all other event types. 
As one can see, with the introduction of fake epochs, \dgem produces higher rates at event occurrences and lower rates at the dead space between occurrences. This results in a sharper intensity rate landscape. Also, it is interesting to observe that some event occurrences lead to the increase of Event Type 1's rate while many others do not. 

\subsection{Graph Visualization of the Attention}
One other advantage of using attention is that it can enable visualization of the relationships among the variables as a graph. We use the average attention of all event channels across time to compute the graph connection.  Specifically, let $A$ be a graph adjacency matrix with element $A_{kq}$, the $k$th row and $q$th column of $A$, indicating event $k$'s influence on the occurrence of event $q$. For clarity of notation, let $\alpha_{ijq}^k = \alpha_{im}^k$ be the computed attention for $k$th channel at time $t_i$. $q \in M$ and $j \in J$ represent the (parental) event type $q$ at time $t_{i-j}$. 

$$ A_{kq} = \frac{1}{T \times J} \sum_{i = 1}^{T} \sum_{j =1}^{J} \alpha_{ijq}^k  $$

\begin{figure}[t]
\centering
\includegraphics[width=0.45\textwidth]{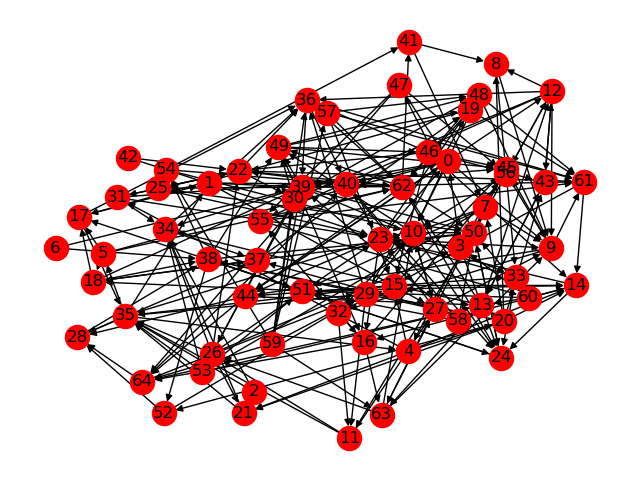}
\caption{Visualization of attention graph in ICEWS Argentina Dataset. }
\label{fig:att}
\end{figure}

We then threshold $ A_{kq}$ to remove small numerical values and obtain $\tilde{A}$. We plot $\tilde{A}$ in Figure~\ref{fig:att} for ICEWS Argentina, where each node is a different event type and edges represent $\tilde{A}$. We use $0.01$ as the threshold. We include the node label in the appendix.  Some example edges learned include: chain 36 -> 40 -> 23, which indicates that when the head of the Argentina government cooperates with the Brazil government, it often leads to internal conflicts between him/her and other branches of the government, and this in turn leads to citizens' unhappiness with the head of the Argentina government. Chain 55 -> 62 indicates that if citizens are in conflict with protestors, citizens will likely cooperate with the police.

\section{Conclusion}

We have introduced a new multi-scale multi-channel neural GEM with two-dimensional attentions for modeling event sequences.  Our model exploits the negative evidence of no observable events in each successive inter-event duration by introducing fake epochs, which eliminates the need to assume specific functional forms. This makes our approach practically appealing with respect to approximately capturing the variation of hidden states in continuous time in a nonparametric manner. Our model combines the framework of GEMs and the modeling power of deep neural networks. On synthetic and benchmark model fitting datasets, our method outperforms other state-of-the-art models by a significant amount, demonstrating that this is a promising approach for modeling event stream data. Alongside the lambda-network in Figure \ref{fig:rnn_basic}, a parallel integral-lambda-network could be used to learn the integral terms in Equation \ref{eqn:log-likelihood} with a  constraint that connects these two networks and their parameters. This would provide an alternative to using a quadrature sum for the integral terms.

The focus of the proposed model with negative evidence is to fit a model for learning dependencies in event stream data. It requires full visibility into the event sequence for the purposes of placement or sampling of the fake epochs. This may not be true in prediction tasks, where the exact use of negative evidence presents an interesting future direction.

\section{Acknowledgement}
We thank Hongyuan Mei from JHU for providing baseline code and results, and anonymous reviewers for their constructive and helpful feedback. 

\bibliographystyle{aaai}
\bibliography{references}

\end{document}